\documentclass[english,english,sort&compress]{article} 
\usepackage[T1]{fontenc}
\usepackage[latin9]{inputenc}
\usepackage{color}
\usepackage{array}
\usepackage{textcomp}
\usepackage{multirow}
\usepackage{amsmath}
\usepackage{amssymb}
\usepackage{graphicx}
\usepackage[numbers]{natbib}

\makeatletter

\providecommand{\tabularnewline}{\\}

\newcommand{\lyxaddress}[1]{
	\par {\raggedright #1
	\vspace{1.4em}
	\noindent\par}
}

\usepackage{orcidlink}
\usepackage{flushend}
\usepackage{arxiv}
\renewcommand{\shorttitle}{Scaling transformation of MMNLSE for PINNs}

\makeatother

\usepackage{babel}
\begin{document}
\title{Scaling transformation of the multimode nonlinear Schr\"odinger equation
for physics-informed neural networks}
\author{\orcidlink{0000-0002-1091-4442}Ivan Chuprov $^{1}$, \orcidlink{0000-0002-7449-5072}Dmitry
Efremenko $^{2}$, \orcidlink{0000-0001-6514-1208}Jiexing Gao $^{1}$,
\orcidlink{0000-0001-6811-2824}Pavel Anisimov $^{1}$, \orcidlink{0000-0002-3932-6759}Viacheslav
Zemlyakov $^{1}$ }
\maketitle

\lyxaddress{$^{1}$ Huawei Technologies Co., Ltd, Central Research Institute,
Shenzhen, China}

\lyxaddress{$^{2}$ Remote Sensing Technology Institute (IMF), German Aerospace
Center (DLR), Oberpfaffenhofen, Germany}
\begin{abstract}
Single-mode optical fibers (SMFs) have become the backbone of modern
communication systems. However, their throughput is expected to reach
its theoretical limit in the nearest future. Utilization of multimode
fibers (MMFs) is considered as one of the most promising solutions
rectifying this capacity crunch. Nevertheless, differential equations
describing light propagation in MMFs are a way more sophisticated
than those for SMFs, which makes numerical modelling of MMF-based
systems computationally demanding and impractical for the most part
of realistic scenarios. Physics-informed neural networks (PINNs) are
known to outperform conventional numerical approaches in various domains
and have been successfully applied to the nonlinear Schr\"odinger
equation (NLSE) describing light propagation in SMFs. A comprehensive
study on application of PINN to the multimode NLSE (MMNLSE) is still
lacking though. To the best of our knowledge, this paper is the first
to deploy the paradigm of PINN for MMNLSE and to demonstrate that
a straightforward implementation of PINNs by analogy with NLSE does
not work out. We pinpoint all issues hindering PINN convergence and
introduce a novel scaling transformation for the zero-order dispersion
coefficient that makes PINN capture all relevant physical effects.
Our simulations reveal good agreement with the split-step Fourier
(SSF) method and extend numerically attainable propagation lengths
up to several hundred meters. All major limitations are also highlighted.
\end{abstract}

\section{Introduction}

Fiber-optic communication systems employ optical fibers for information
transfer. The state-of-the-art fiber-optic communication systems have
advanced dramatically in the last two decades. The so called bit rate-distance
product $BL$, where $B$ and $L$ stand for the bit rate and the
repeater spacing, respectively, has increased by three orders of magnitude
in the last decade \citep{Winzer:15,Agrawal2021}. However, ever-growing
information traffic demands enforced reconsideration of future optical
communication technology and boosted research on novel transmission
approaches, such as mode division multiplexing (MDM) \citep{Puttnam:21}.
Opposed to current single-mode fibers (SMFs) systems, MDM implies
utilization of multimode fibers (MMFs) that support simultaneous propagation
of multiple guided modes. These modes, in turn, can be encoded as
independent signals and thus enable spatial multiplexing. Despite
such systems have already surpassed the important milestone of 1 Pb/s
data throughput \citep{Rademacher2021}, MDM is still in its infancy.
It is largely because of additional descriptive complexity of light
propagation in MMFs that is mostly associated with intertwinement
of different modes. Namely, conventional single-mode nonlinear Schr\"odinger
equation (NLSE) is replaced by a system of coupled differential equations,
referred to as multimode NLSE (MMNLSE). Ubiquitous tools for solving
NLSE, such as the split-step Fourier method (SSF) \citep{Sinkin2003,Shao2014},
are naturally extendable to the case of MMNLSE but require tremendous
computational effort, whereas realistic MMF systems of 200-500 m \citep{ADDANKI2018743}
cannot be decently modeled without sacrificing MMNLSE generality.
This fact gives rise to new challenges that numerical algorithms have
to address. Such algorithms are required to model accurately the signal
propagation in fibers and, for example, to design techniques for compensating
signal distortions \citep{Li21}.

Recently, machine learning (ML) techniques were applied to NLSE \citep{Jiang2022}.
One of the most promising ML tools for solving NLSE is physics-informed
neural networks (PINNs). PINNs belong to universal function approximators
that are trained with an eye on the underlying physical laws rather
than given on datasets \citep{raissi2017physicsI}. In practice, PINNs
are used to solve partial differential equations (PDEs) and in this
regard, are often considered as a mesh-free alternative to traditional
numerical solvers. The idea of the PINN approach consists in training
a neural network so that it satisfies both a given PDE and boundary/initial
conditions. Consequently, the loss function to be minimized during
the training process consists of the residual w.r.t. the PDE (i.e.
when substituting the PINN into the original PDE) and the residual
w.r.t. the initial conditions. The key feature of the PINNs approach
is that the corresponding partial derivatives of the neural network
are computed with help of automatic differentiation tools implemented
in ML libraries (e.g. PyTorch). In this manner, the loss function
can be minimized and the given PDE can be solved using the tools of
ML in a weak formulation. It can be shown that PINNs methodology is
similar to Newton-Krylov solvers \citep{Markidis2021,Kelley1995}
and the finite element method \citep{Reddy2018}. In addition, PINNs
are conceptually similar to the Ritz method \citep{Weinan2018}.

PINNs were applied to a single-mode NLSE and validated against the
SSF \citep{Monterola2001,Monterola2003,9489953,raissi2019physics}.
Although PINNs can provide accurate results, their current training
time is larger than that of SSF. Therefore, PINNs cannot compete with
the classical solvers. However, some features of PINNs motivate their
further development resulting in a large amount of modifications (see
\citep{Shin2020,Karniadakis2020,Jagtap2020,Yu2022} and references
therein). The first advantage of PINNs is their universality, as they
have been applied to problems in the wide range of fields \citep{Sun2020,Sahli_Costabal2020,Zhang2020}.
Secondly, PINNs have the potential to be an effective analysis tool
for long-distance and high-channel loading scenarios, as its complexity
raises slower with the pulse energy than that of SSF \citep{9489953}.
Thirdly, the training time of PINNs can be significantly reduced via
transfer-learning. It has been estimated that the total number of
peer-reviewed papers about PINNs applications has exceeded 2000, and
this quantity has grown from 30 papers in 2018 to 1300 in 2021 \citep{Cuomo2022}
.

The main drawback of PINNs (apart from the current long training time)
is that they may fail to converge in some cases depending on the parameters
entering a PDE. It was noted that the convergence of PINN for solving
NLSE is much slower than that for the linear Schr\"odinger equation,
i.e. more time is required to capture nonlinear dependencies \citep{Monterola2003}.
Nevertheless, such an observation does not explain why in some cases
PINNs yield no solution at all. An attempt to study these PINNs issues
from the mathematical point of view was done in \citep{wang2021and}.
Besides, PINNs were improved by adding new modifications in the neural
network \citep{Wang2021,wang2021and,Yu2022}. However, these approaches
are not suitable for MMNLSE, since the PINNs fail to converge, and
to the best of our knowledge, no such results have been reported so
far. 

In this paper, we propose a recipe for solving MMNLSE with PINN. Considering
this problem from the physical point of view, we use a scale transformation
of MMNLSE parameters, which allows leveling a large difference between
coefficients and leads to a successful solution for a hundred of meters.
We also note that it is necessary to consider the maximum of fiber
length for PINN only with a energy value. We demonstrate that the
energy value affects the nonlinear length which constrains PINNs.

This paper is organized as follows. In Section \ref{sec:PINN-configuration},
we introduce a suitable PINN configuration. In Section \ref{sec:Theory},
NLSE is reviewed and the origin of numerical problems for PINNs is
discussed. A frame transformation and normalization procedure are
analyzed and the scaling transformation is proposed which allows us
to bring MMNLSE to the PINN solvable form. Section \ref{sec:Numerical-results}
contains results of numerical experiments and provide comparison of
the PINN with SSF. The paper is concluded with a summary.

\section{PINN configuration\label{sec:PINN-configuration}}

\begin{figure}
\begin{centering}
\includegraphics[width=0.8\columnwidth]{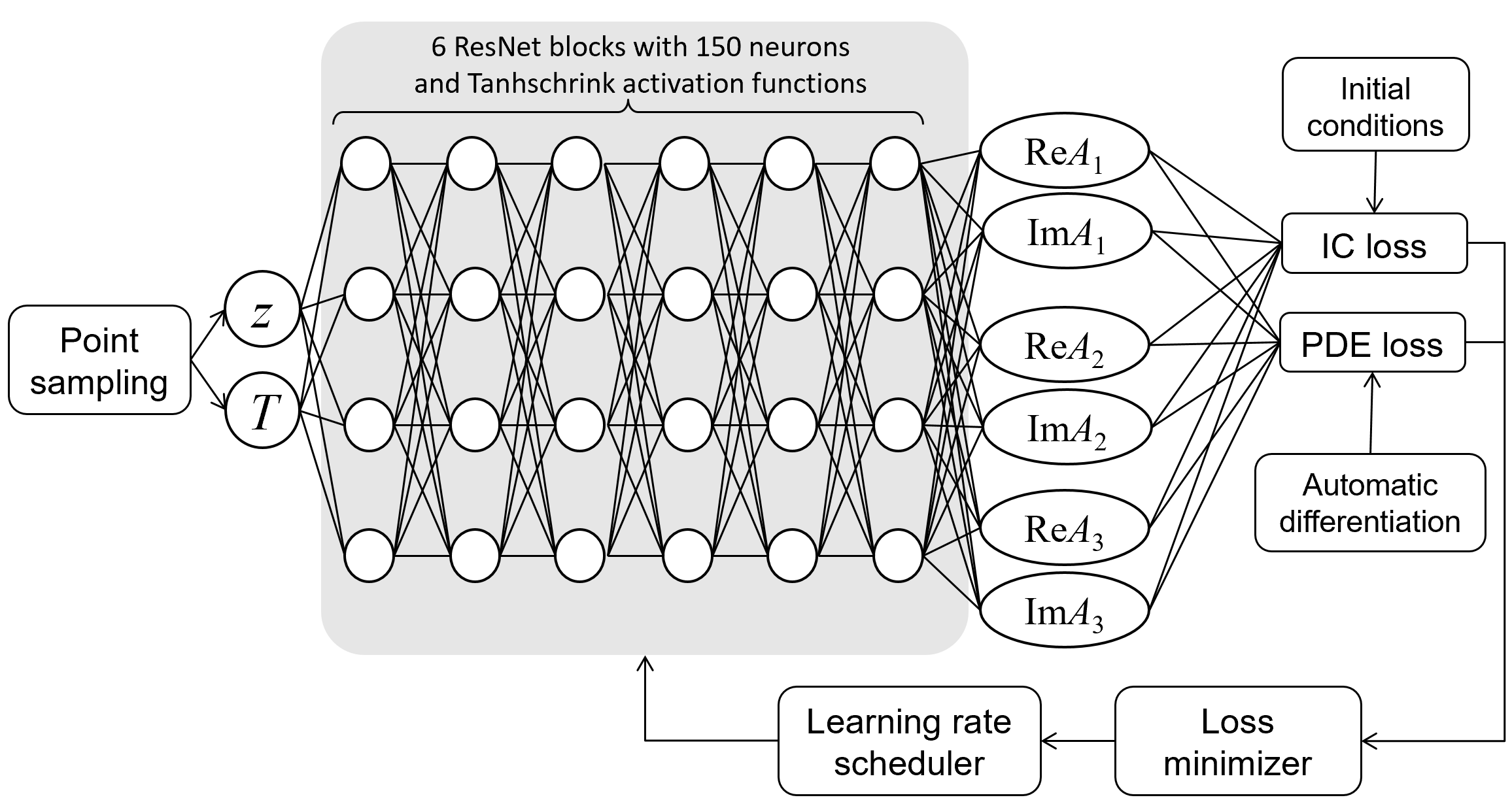}
\par\end{centering}
\caption{PINN configuration used in this study}
\label{pinn_scheme1}
\end{figure}

We developed a PINN framework using the PyTorch library \citep{NEURIPS2019_9015}
as a backend. The neural network configuration is organized as follows.
The PINN model consists of 6 ResNet blocks with 150 neurons and Tanhshrink
activation functions in each of them. This configuration is chosen
empirically: other numbers of blocks, neurons, or changes in other
parameters worsened our results. In our simulations we also observed
that Tanhshrink performs better than ReLU and Tanh activation functions.
The actual value of the learning rate is governed by the scheduler.
It multiplies the current value of the learning rate by factor (0.9
by default) if the loss function is not reduced after a certain amount
of iterations (30 iterations by default). The minimal learning rate
value is set to 1e-7. The partial derivatives are computed by using
built-in features of PyTorch automatic differentiation. The PINN takes
the pairs of $T$ and $z$ as input and returns real and imaginary
components for each mode. Thus, PINN has 2 inputs and 6 outputs. The
PINN structure is summarized in Fig. \ref{pinn_scheme1}. The training
is run on GPU NVidia Tesla V100 16GB.

240000 sample points are mainly used. We have found that this is sufficient
to obtain satisfactory results in our cases. Empirically it was found
that by taking more sampling points around $T=0$ (i.e. where the
function changes faster), PINN accuracy can be improved. Keeping that
in mind, we adopt a heterogeneous algorithm for random point sampling
which provides 90\% of points in the time domain corridor $[-T_{\max}/2,T_{\max}/2]$.
The weights of the network are initialized by using the Xavier method
\citep{pmlr-v9-glorot10a}. 

Note that the loss function is not a straightforward measure of the
PINN accuracy. Therefore, for validation purposes, the PINN results
are compared with the SSF solution. Optionally, frame transformation
(\ref{eq:k1})-(\ref{eq:k2}), pulse normalization (\ref{eq:norm})
as well as the novel scaling transformation (\ref{eq:bet})-(\ref{eq:bet1})
for $\delta\beta_{0}^{(p)}$ can be enabled.

\section{Theory\label{sec:Theory}}

\subsection{Problem background}

\begin{figure}
\begin{centering}
\includegraphics[width=0.6\columnwidth]{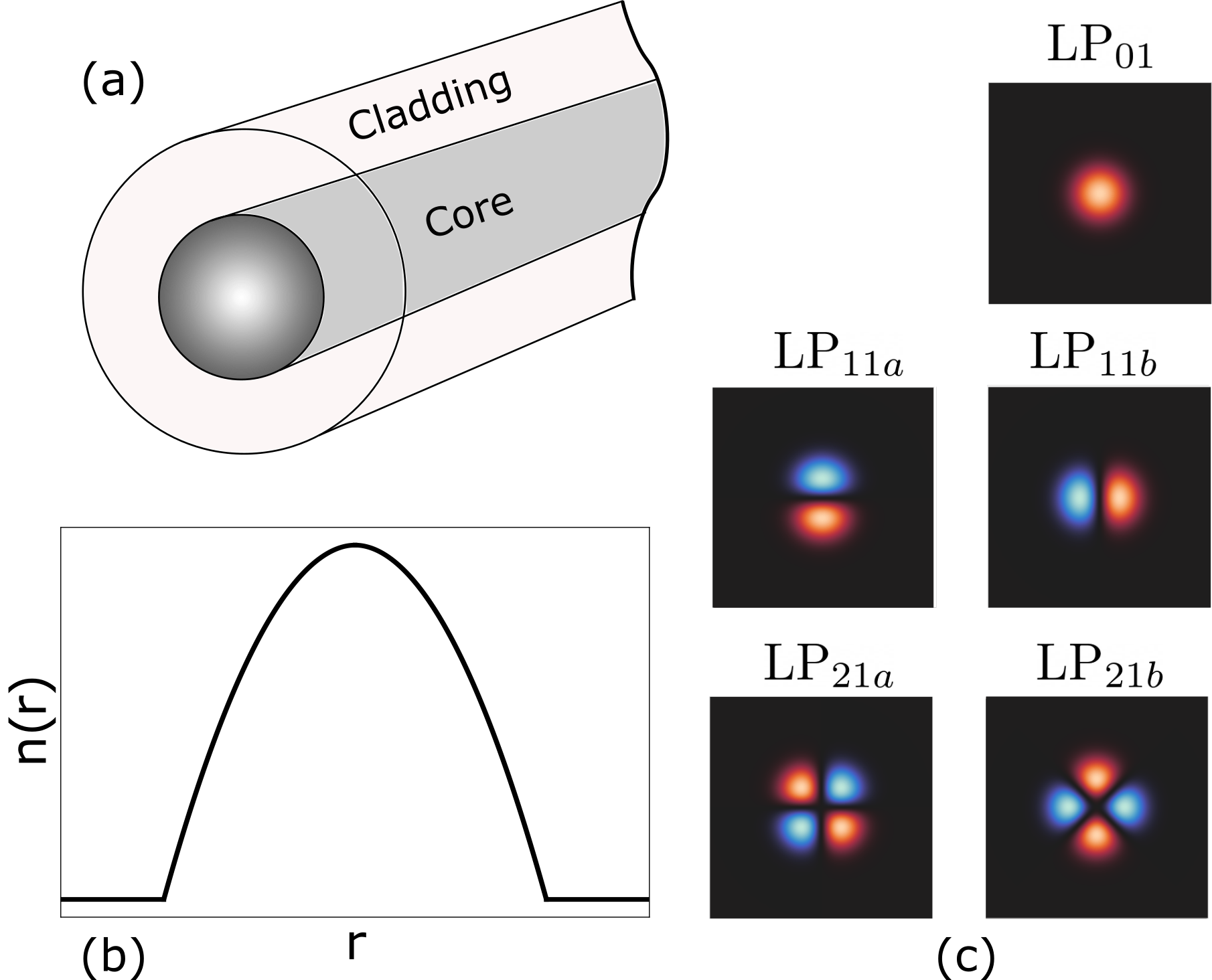}
\par\end{centering}
\caption{Modes in multimode fiber. (a) Schematic of a graded-index multimode
fiber; (b) GRIN profile; (c) Intensity profiles of first 5 modes.}

\label{fig:modes}
\end{figure}

Optical fibers consist of a core surrounded by a cladding. The former
has higher values of the refractive index that results in total internal
reflection, and thus confines light (Fig. \ref{fig:modes}(a)). Light
propagation through an MMF is derived from the Maxwell's equation,
which is solved by separation of variables. The transversal part satisfies
the Helmholtz equation that in the case of large core diameters enables
multiple solutions, referred to as modes. The fiber, in this case,
is called an MMF. Modal content of such optical fibers strongly depends
on the shape of the refractive index profile function (Fig. \ref{fig:modes}(b))
that describes its evolution along the radial direction. If $n_{core}\backsimeq n_{clad}$,
the so-called weakly-guided approximation takes place and we call
the resulting solution the linearly-polarized modes (LP). These are
classified by two indices originating from the Helmhpolz equation
and are ordered in the descending order with respect to their propagation
constants determining phase velocities of the modes \citep{Agrawal2021}.
If the shape of the refractive index profile function fulfills a quasi-parabolic
law (the so-called graded-index profile function or GRIN, Fig. \ref{fig:modes}(b)),
modes become degenerate with respect to the propagation constants
and reveal similar propagation behaviour. 

In this work, we consider propagation of the first three LP modes
implying that degenerate partners are omitted, i.e. LP$_{01}$, LP$_{11a}$
and LP$_{21a}$ (Fig. \ref{fig:modes}(c)). In what follows, we directly
address them by their appearence order as 1,2 and 3.

\subsection{MMNLSE and the frame transformation}

Evolution of a laser pulse in MMF is described by MMNLSE \citep{Agrawal2021}

\begin{align}
\mathfrak{N}(A_{p})\equiv & -\dfrac{\partial A_{p}}{\partial z}+i\delta\beta_{0}^{(p)}A_{p}-i\delta\beta_{1}^{(p)}\dfrac{\partial A_{p}}{\partial T}-i\dfrac{\beta_{2}^{(p)}}{2}\dfrac{\partial^{2}}{\partial T^{2}}A_{p}\label{eq:linear_multi}\\
 & +i\gamma_{S}\left|A_{p}\right|^{2}A_{p}+i\sum_{n\neq p}\gamma_{C}^{(n)}\left|A_{n}\right|^{2}A_{p}=0,\nonumber 
\end{align}
where $A(z,T)$ is the electric field temporal envelope, $z$ is the
propagation distance, $\delta\beta_{j}^{(p)}=\beta_{j}^{(p)}-{\textstyle \text{Re}}\beta_{j}^{(0)}$,
$\beta_{j}^{(p)}$ is the $j$th order dispersion coefficient for
the $p$ th mode, $T$ is the time under time-retarded frame, while
$\gamma_{S}$ and $\gamma_{C}^{(n)}$ are the nonlinear coefficients,
namely self-phase and cross-phase modulations Kerr coefficients, respectively
\citep{Agrawal2021}. Nonlinearity is mainly responsible for coupling
between modes and leads to various unexpected effects and problems.
In this paper, four wave mixing, Raman scattering and the fiber losses
are neglected.

During the learning process, PINN is substituted into Eq. (\ref{eq:linear_multi})
as $A_{p}$ and partial derivatives are computed by automatic differentiation
tools. The PDE residual is then equal to $\mathfrak{N}(\text{PINN})$
which is to be minimized together with the initial condition residual.
Note if the coefficients in Eq. (\ref{eq:linear_multi}) differ by
several orders of magnitude, the PINN becomes less sensible to the
terms with small ones, and, therefore, their influence cannot be captured. 

It is a common practice to consider $A$ as a function of unitless
variables $\zeta$ and $t$ defined through the following relations
\citep{normalization}

\begin{equation}
z=k_{1}L_{D}\zeta,\:T=k_{2}T_{0}t,\label{eq:k1}
\end{equation}

where 

\begin{equation}
k_{1}=\dfrac{L_{\max}}{L_{D}},\:k_{2}=\dfrac{T_{\max}}{T_{0}},\label{eq:k2}
\end{equation}
$L_{D}$ is the dispersion length given by

\begin{equation}
L_{D}=\dfrac{T_{0}^{2}}{\left|\beta_{2}\right|},
\end{equation}
with $[-T_{max}/2$, $T_{max}/2]$ being the time window and $T_{0}$
the initial width of the pulse. Eqs. (\ref{eq:k1})-(\ref{eq:k2})
are referred to as frame transformation. In addition, $A$ can be
normalized by the peak power of the pulse $P_{0},$ 

\begin{equation}
A=\sqrt{P_{0}}U.\label{eq:norm}
\end{equation}
Applying transformations (\ref{eq:k1})-(\ref{eq:k2}) and (\ref{eq:norm})
to Eq. (\ref{eq:linear_multi}), we obtain the MMNLSE for the normalized
quantity $U$ in the transformed frame $(\xi,t)$

\begin{align}
 & i\dfrac{\partial U_{p}}{\partial\zeta}+k_{1}\delta\beta_{0}^{(p)}U_{p}+\dfrac{k_{1}\delta\beta_{1}^{(p)}}{k_{2}}\dfrac{\partial U_{p}}{\partial t}+\dfrac{k_{1}c}{k_{2}^{2}}\dfrac{\partial^{2}U_{p}}{\partial t^{2}}\label{eq:linear_multi-1}\\
 & +k_{1}d\left|U_{p}\right|^{2}U_{p}+k_{1}d\sum_{n\neq p}\dfrac{\gamma_{C}^{(n)}}{\gamma_{S}}\left|U_{n}\right|^{2}U_{p}=0,\nonumber 
\end{align}
where
\begin{equation}
c=\dfrac{-\mathrm{sign}(\beta_{2})}{2},\:d=\dfrac{L_{D}}{L_{NL}},
\end{equation}
while

\begin{equation}
L_{NL}=\dfrac{1}{\gamma P_{0}}\label{eq:nl-2}
\end{equation}
is the so-called nonlinear length. Coefficients used in this study
are given in Tables \ref{tab:coefs}, \ref{tab:coefs1} for Eq. (\ref{eq:linear_multi})
and (\ref{eq:linear_multi-1}) respectively. Examples of coefficients
for cases considered in this study are shown in Table for Eq. and
Table for Eq. (\ref{eq:linear_multi}). Note, the coefficients in
Eq. (\ref{eq:linear_multi}) depend on the fiber material and the
pulse wavelength, while the weights at the terms of Eq. (\ref{eq:linear_multi-1})
additionally depend on the pulse energy and the time window. It can
be seen that transformations (\ref{eq:k1})-(\ref{eq:k2}) and (\ref{eq:norm})
change the order of certain coefficients, which appear to be beneficial
for some cases (e.g. energies $\sim1{}^{-6}$ nJ). However, these
transformations do not solve the problem associated with large values
of $\delta\beta_{0}^{(p)}$ (as compared to other coefficients\textcolor{red}{{}
}in Eq. (\ref{eq:linear_multi-1})), but rather increase the difference
between them.  From the physical point of view, high values of $\delta\beta_{0}^{(p)}$
cause the high-frequency dependency of $U_{p}$ with respect to $z$,
thus creating a problem for PINNs. Our goal is to reduce the large
values of the $\delta\beta_{0}^{(p)}$ coefficients in a reasonable
way. We consider the Schr\"odinger equation step by step, adding
to it multimodality and nonlinearity. 

\begin{table*}
\caption{Coefficients of Eq. (\ref{eq:linear_multi})}
\label{tab:coefs}
\centering{}%
\begin{tabular}{ccccccc}
\hline 
\multirow{4}{*}{$\lambda$, nm} & \multicolumn{6}{c}{Terms}\tabularnewline
\cline{2-7} \cline{3-7} \cline{4-7} \cline{5-7} \cline{6-7} \cline{7-7} 
 & $\partial A_{p}/\partial z$ & $A_{p}$ & $\partial A_{p}/\partial t$ & $\partial^{2}A_{p}/\partial t^{2}$ & $\left|A_{p}\right|^{2}A_{p}$ & $\left|A_{n}\right|^{2}A_{p}$\tabularnewline
\cline{2-7} \cline{3-7} \cline{4-7} \cline{5-7} \cline{6-7} \cline{7-7} 
 & \multicolumn{6}{c}{Coefficients}\tabularnewline
\cline{2-7} \cline{3-7} \cline{4-7} \cline{5-7} \cline{6-7} \cline{7-7} 
 & 1 & $\delta\beta_{0}^{(p)}$, m$^{-1}$ & $\delta\beta_{1}^{(p)}$, ps/m & $\beta_{2}^{(p)}/2$, ps$^{2}$/m & $\gamma_{S}$, W$^{-1}$/m & $\gamma_{C}^{(n)}$, W$^{-1}$/m$^{*}$\tabularnewline
\hline 
\hline 
1030 & 1 & --11328 & 0.0079 & 0.0096 & 0.0011 & 0.0006\tabularnewline
\hline 
\multicolumn{7}{l}{$^{*}$ Specified maximum coefficient values for all modes}\tabularnewline
\end{tabular}
\end{table*}

\begin{table*}

\caption{Coefficients of Eq. (\ref{eq:linear_multi-1}) for different values
of pulse energy $E$ and fiber length $L$}

\label{tab:coefs1}
\centering{}%
\begin{tabular}{ccccccccccc}
\hline 
\multirow{4}{*}{$E$, nJ} & \multirow{4}{*}{$P_{0}$, W} & \multirow{4}{*}{$L$, m} & \multirow{4}{*}{$L_{D}$, m} & \multirow{4}{*}{$L_{NL}$, m} & \multicolumn{6}{c}{Terms}\tabularnewline
\cline{6-11} \cline{7-11} \cline{8-11} \cline{9-11} \cline{10-11} \cline{11-11} 
 &  &  &  &  & $\partial U_{p}/\partial\zeta$ & $U_{p}$ & $\partial U_{p}/\partial t$ & $\partial^{2}U_{p}/\partial t^{2}$ & $\left|U_{p}\right|^{2}U_{p}$ & $\left|U_{n}\right|^{2}U_{p}$\tabularnewline
\cline{6-11} \cline{7-11} \cline{8-11} \cline{9-11} \cline{10-11} \cline{11-11} 
 &  &  &  &  & \multicolumn{6}{c}{Coefficients}\tabularnewline
\cline{6-11} \cline{7-11} \cline{8-11} \cline{9-11} \cline{10-11} \cline{11-11} 
 &  &  &  &  & 1 & $k_{1}\delta\beta_{0}^{(p)}$ & $k_{1}\delta\beta_{1}^{(p)}/k_{2}$ & $k_{1}c/k_{2}^{2}$ & $k_{1}d$ & $k_{1}d(\gamma_{C}^{(n)}/\gamma_{S})$$^{*}$\tabularnewline
\hline 
\hline 
10 & 9393 & 5 & 18.8 & 0.09 & 1 & -3009 & 1.3e-5 & -4.8e-6 & 53.85 & 107.7\tabularnewline
\hline 
0.1 & 93 & 100 & 18.8 & 9.28 & 1 & -60182 & 2.5e-4 & -9.6e-5 & 10.77 & 21.54\tabularnewline
\hline 
1e-3 & 0.93 & 1000 & 18.8 & 928 & 1 & -601829 & 8.4e-4 & -1e-4 & 1.077 & 2.154\tabularnewline
\hline 
1e-6 & 9.3e-3 & 1000 & 18.8 & 928475 & 1 & -601829 & 8.4e-4 & -1e-4 & 1e-3 & 2.1e-3\tabularnewline
\hline 
\multicolumn{11}{l}{$^{*}$ Specified maximum coefficient values for all modes}\tabularnewline
\end{tabular}
\end{table*}

\subsection{Single-mode linear case}

We start with a single-mode linear equation and revise the derivation
of its solution. The linear single mode equation for $A(z,T)$ reads
as follows:

\begin{equation}
i\dfrac{\partial A}{\partial z}=\dfrac{\beta_{2}}{2}\dfrac{\partial^{2}A}{\partial T^{2}}.\label{eq:linear_signl}
\end{equation}
The initial condition is given by the Gaussian pulse 

\begin{equation}
A(0,T)=\exp\left\{ -\frac{T^{2}}{2T_{0}^{2}}\right\} ,\label{eq:bound-1}
\end{equation}
where $T_{0}$ is the half-width, that related to the full-width at
half-maximum (FWHM) of the pulse as

\begin{equation}
T_{FWHM}=2(\mathrm{ln}2)^{1/2}T_{0}\approx1.665T_{0}.
\end{equation}
Following \citep{Agrawal2021}, we can analytically solve Eq. (\ref{eq:linear_signl})
(for details see Appendix 1). The final result reads as

\begin{equation}
A\left(z,T\right)=\dfrac{T_{0}}{\sqrt{T_{0}^{2}-i\beta_{2}z}}\exp\left[-\dfrac{T^{2}}{2\left(T_{0}^{2}-i\beta_{2}z\right)}\right].\label{eq:lin}
\end{equation}

\subsection{Multimode linear case}

Now we consider the linear equation with the additional term $\delta\beta_{0}^{(p)}A$
which is present in the multimode equation

\begin{equation}
i\dfrac{\partial A}{\partial z}=\dfrac{\beta_{2}}{2}\dfrac{\partial^{2}A}{\partial T^{2}}+\delta\beta_{0}^{(p)}A.
\end{equation}
Applying Fourier transform we obtain

\begin{align}
i\dfrac{\partial\hat{A}}{\partial z}= & -\dfrac{1}{2}\beta_{2}\omega^{2}\hat{A}+\delta\beta_{0}^{(p)}\hat{A}\\
= & \left(-\dfrac{1}{2}\beta_{2}\omega^{2}+\delta\beta_{0}^{(p)}\right)\hat{A}.\label{eq:MM fourier}
\end{align}
Solution of. Eq (\ref{eq:MM fourier}) reads as follows

\begin{equation}
\hat{A}\left(z,\omega\right)=\hat{A}\left(0,\omega\right)\exp\left\{ \left(\dfrac{i}{2}\beta_{2}\omega^{2}-i\delta\beta_{0}^{(p)}\right)z\right\} .
\end{equation}
The inverse Fourier transform yields

\begin{align}
A\left(z,T\right)=\dfrac{1}{2\pi} & \int_{-\infty}^{+\infty}\hat{A}\left(0,\omega\right)\\
 & \times\exp\left\{ \left(\dfrac{i}{2}\beta_{2}\omega^{2}-i\delta\beta_{0}^{(p)}\right)z-i\omega T\right\} d\omega.\nonumber 
\end{align}
Using the initial condition (\ref{eq:bound-1}), we obtain

\begin{align}
A\left(z,T\right)=\dfrac{1}{2\pi} & \int_{-\infty}^{+\infty}T_{0}\exp\left\{ -\frac{1}{2}T_{0}^{2}\omega^{2}\right\} \\
 & \times\exp\left\{ \left(\dfrac{i}{2}\beta_{2}\omega^{2}-i\delta\beta_{0}^{(p)}\right)z-i\omega T\right\} d\omega.\nonumber 
\end{align}
The term $\exp\left\{ -i\delta\beta_{0}^{(p)}z\right\} $ can be moved
outside the integral

\begin{align}
A\left(z,T\right)=\exp\left\{ -i\delta\beta_{0}^{(p)}z\right\} \dfrac{1}{2\pi} & \int_{-\infty}^{+\infty}T_{0}\exp\left\{ -\frac{1}{2}T_{0}^{2}\omega^{2}\right\} \\
 & \times\exp\left\{ \dfrac{i}{2}\beta_{2}\omega^{2}z-i\omega T\right\} d\omega.\nonumber 
\end{align}
The final result reads as follows

\begin{align}
A\left(z,T\right)= & \exp\left\{ -i\delta\beta_{0}^{(p)}z\right\} \dfrac{T_{0}}{\sqrt{T_{0}^{2}-i\beta_{2}z}}\label{eq:lin1}\\
 & \times\exp\left\{ -\dfrac{T^{2}}{2\left(T_{0}^{2}-i\beta_{2}z\right)}\right\} .\nonumber 
\end{align}
Recalling that $A$ is a function of $A\left(z,T,\delta\beta_{0}^{(p)}\right)$,
with help of Eqs. (\ref{eq:lin}) and (\ref{eq:lin1}) we obtain

\begin{equation}
A\left(z,T,\delta\beta_{0}^{(p)}\right)=\exp\left\{ -i\delta\beta_{0}^{(p)}z\right\} A\left(z,T,0\right).\label{eq:11}
\end{equation}
Consequently, $A\left(z,T,\delta\beta_{0}^{(p)}\right)$ is periodic
in $\delta\beta_{0}^{(p)}$ with period $\dfrac{2\pi}{z}$

\begin{equation}
A\left(z,T,\delta\beta_{0}^{(p)}\right)=A\left(z,T,\delta\beta_{0}^{(p)}+\dfrac{2\pi}{z}l\right),\label{eq:slaing}
\end{equation}
where $l$ is integer.

\subsection{Nonlinear case: self-phase and cross-phase modulations}

We examine whether the same periodicity is valid for nonlinear cases
and consider NLSE with $\delta\beta_{0}^{(p)}A$ term

\begin{equation}
i\dfrac{\partial A}{\partial z}=\dfrac{\beta_{2}}{2}\dfrac{\partial^{2}A}{\partial T^{2}}+\delta\beta_{0}^{(p)}A+\gamma_{S}\left|A\right|^{2}A.\label{eq:nl}
\end{equation}
In addition, we focus on

\begin{equation}
i\dfrac{\partial\bar{A}}{\partial z}=\dfrac{\beta_{2}}{2}\dfrac{\partial^{2}\bar{A}}{\partial T^{2}}+\gamma_{S}\left|\bar{A}\right|^{2}\bar{A}.\label{eq:qq}
\end{equation}
Similar to the previous subsection, if we show that
\begin{equation}
A(z,T)=\bar{A}(z,T)\exp\left\{ -i\delta\beta_{0}^{(p)}z\right\} ,\label{eq:anz}
\end{equation}
then the periodicity with respect to $\beta_{0}$ is proven automatically.
Substitution of Eq. (\ref{eq:anz}) into Eq. (\ref{eq:nl}) yields

\begin{align*}
 & i\dfrac{\partial\bar{A}}{\partial z}\exp\left\{ -i\delta\beta_{0}^{(p)}z\right\} +\delta\beta_{0}^{(p)}\bar{A}\exp\left\{ -i\delta\beta_{0}^{(p)}z\right\} \\
 & =\dfrac{\beta_{2}}{2}\dfrac{\partial^{2}\bar{A}}{\partial T^{2}}\exp\left\{ -i\delta\beta_{0}^{(p)}z\right\} +\delta\beta_{0}^{(p)}\bar{A}\exp\left\{ -i\delta\beta_{0}^{(p)}z\right\} \\
 & +\gamma_{S}\left|\bar{A}\right|^{2}\bar{A}\exp\left\{ -i\delta\beta_{0}^{(p)}z\right\} .
\end{align*}
By canceling $\exp\left\{ -i\delta\beta_{0}^{(p)}z\right\} $ and
taking into account Eq. (\ref{eq:qq}) we obtain an identity. Thus,
the periodicity property (\ref{eq:slaing}) also hold for Eq. (\ref{eq:nl}).
Note, in the case of self-phase modulation, the equations for the
modes do not depend on each other and can be solved by a PINN independently.

Finally, we consider NLSE for $A\equiv A_{p}$ with the $\delta\beta_{0}^{(p)}A$
term and the term $\gamma_{C}\left|A_{n}\right|^{2}A$, $n\neq p$,
which governs the cross-phase modulation

\begin{equation}
i\dfrac{\partial A}{\partial z}=\dfrac{\beta_{2}}{2}\dfrac{\partial^{2}A}{\partial T^{2}}+\delta\beta_{0}^{(p)}A+\gamma_{C}\left|A_{n}\right|^{2}A.\label{eq:nl-1}
\end{equation}
It is worth mentioning that the periodic property of $A_{n}$ (see
Eq. (\ref{eq:slaing})) does not affect $A$ in Eq. (\ref{eq:nl-1}),
whereas the rest of the proof is analogous to that of the self-phase
modulation case. Therefore, Eq. (\ref{eq:slaing}) is valid for cross-phase
modulation as well.

\subsection{Scale transformation }

\begin{figure}
\begin{centering}
\includegraphics[width=0.8\columnwidth]{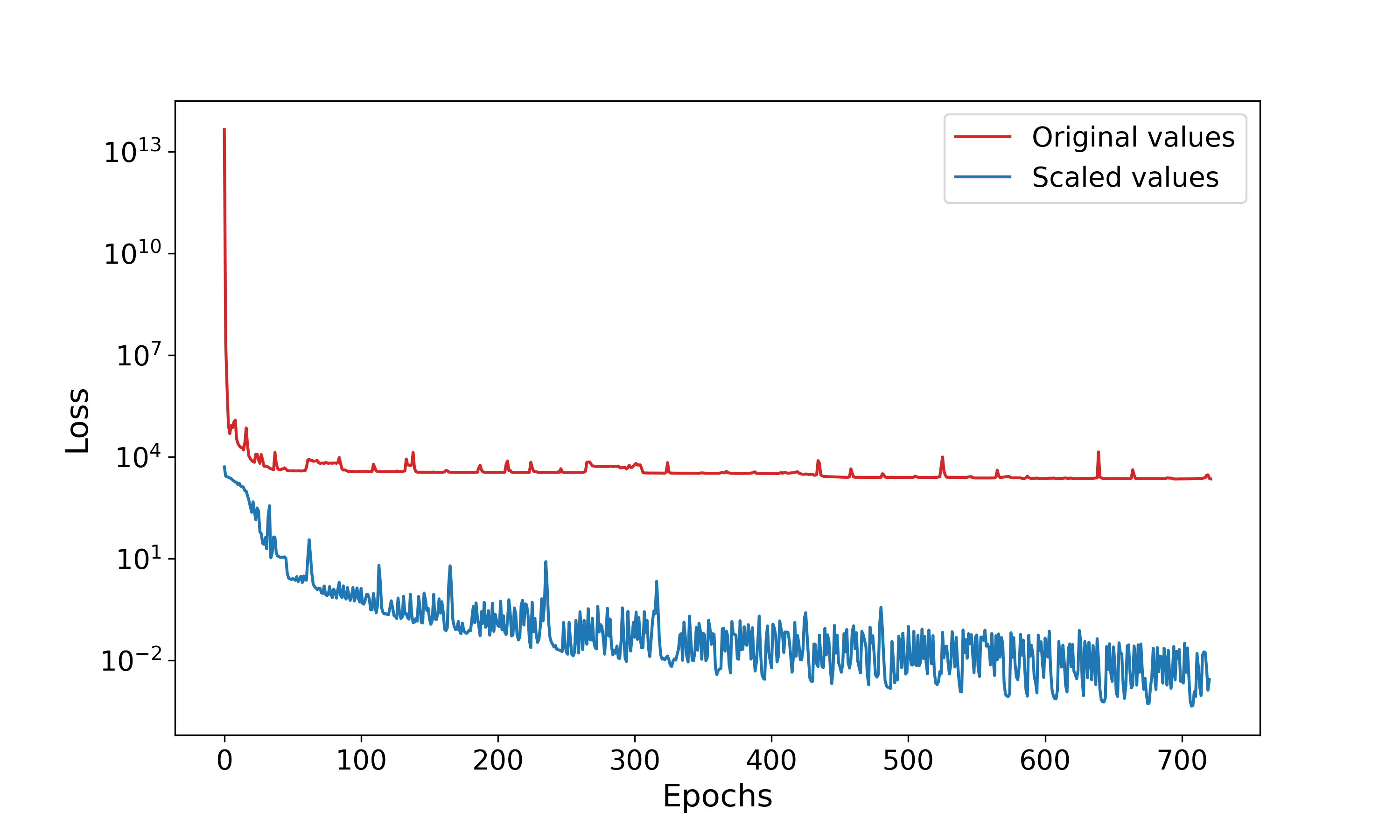}
\par\end{centering}
\caption{Comparison of the loss functions for the original and scaled values
of $\delta\beta_{0}^{(p)}$}

\label{fig:loss}
\end{figure}

Recalling the property (\ref{eq:slaing}), we conclude that the solution
at the coordinate $z$ does not change if we replace $\delta\beta_{0}^{(p)}$with

\begin{equation}
\widehat{\delta\beta_{0}^{(p)}}=\delta\beta_{0}^{(p)}+\dfrac{2\pi}{z}l.
\end{equation}
Since our goal is to reduce the absolute value of the coefficient
by the $\delta\beta_{0}^{(p)}A$ term we can set

\begin{equation}
\widehat{\delta\beta_{0}^{(p)}}=\delta\beta_{0}^{(p)}+\dfrac{2\pi}{z}l',\label{eq:bet}
\end{equation}
where 

\begin{equation}
l'=\left\lfloor \dfrac{\delta\beta_{0}^{(p)}z}{2\pi}\right\rfloor ,\label{eq:bet1}
\end{equation}
and the brackets '$\left\lfloor .\right\rfloor $' stand for the integer
part. Accroding to Eq. (\ref{eq:11}), for a point $x$ inside the
fiber ($0<x<z$) holds 

\begin{align}
 & A\left(x,T,\delta\beta_{0}^{(p)}+\dfrac{2\pi}{z}l\right)\\
 & =\exp\left\{ -i\left(\delta\beta_{0}^{(p)}x+\dfrac{2\pi}{z}lx\right)\right\} A\left(x,T,0\right).\nonumber 
\end{align}
Thus, the scale transformation (\ref{eq:bet}) for the $\delta\beta_{0}^{(p)}A$
term affects neither the absolute value of $A_{p}$ across the entire
fiber nor its phase expressed via $\mathrm{Re}A_{p}$ and $\mathrm{Im}A_{p}$
at points $x,$ such that $lx/z$ is integer. With help of this scale
transformation, we can level the order of the MMNLSE coefficients,
thus overcoming the obstacle that prevents PINN from convergence.
The reference $\delta\beta_{0}^{(p)}$ and scaled values $\widehat{\delta\beta_{0}^{(p)}}$
for different lengths of fiber are summarized in Table \ref{tab:tab_b}.
As a result of application of the scaled values $\widehat{\delta\beta_{0}^{(p)}}$
, PINN converges, as evidenced by the decreasing loss function. Comparison
of loss functions for the original and the scaled values is presented
in Fig. \ref{fig:loss}. Despite the fact, that the absolute value
of $A$ does not depend on $\delta\beta_{0}^{(p)}$, PINN converges
neither for real and imaginary parts of $A$, nor the absolute value
of $A.$ We can see that in th case of unscaled dispersion, the loss
function starts with very large values and sharply  reaches a plateau
about values $\sim10^{4}$. If the dispersion coefficients are scaled,
the loss function starts from $\sim10^{4}$ and continues to decrease
throughout the training period.

\begin{table}
\caption{Reference and scaled values of $\delta\beta_{0}^{(p)}$}
\label{tab:tab_b}
\centering{}%
\begin{tabular}{ccccc}
\hline 
$p$ & $\delta\beta_{0}^{(p)}$, m$^{-1}$ & $\widehat{\delta\beta_{0}^{(p)}}$, m$^{-1}$ & $\left\lfloor \delta\beta_{0}^{(p)}z/(2\pi)\right\rfloor ${*} & $L$, m\tabularnewline
\hline 
\hline 
1 & 0 & --- & --- & ---\tabularnewline
\hline 
\multirow{2}{*}{2} & \multirow{2}{*}{-5662.25183} & -0.09655 & 90116 & 100\tabularnewline
\cline{3-5} \cline{4-5} \cline{5-5} 
 &  & -1.10186 & 4505 & 5\tabularnewline
\hline 
\multirow{2}{*}{3} & \multirow{2}{*}{-11328.0841} & -0.0036 & 180292 & 100\tabularnewline
\cline{3-5} \cline{4-5} \cline{5-5} 
 &  & -0.7576 & 9014 & 5\tabularnewline
\hline 
\multicolumn{5}{l}{{*} Integer number of periods}\tabularnewline
\end{tabular}
\end{table}

Alternatively, recalling Eq. (\ref{eq:11}) we have

\begin{align}
A\left(z,T,\delta\beta_{0}^{(p)}\right)= & \left(\cos\left\{ -\delta\beta_{0}^{(p)}z\right\} +i\sin\left\{ -\delta\beta_{0}^{(p)}z\right\} \right)\\
 & \times\left(\mathrm{Re}A\left(z,T,0\right)+i\mathrm{Im}A\left(z,T,0\right)\right),\nonumber 
\end{align}
which yields

\begin{align}
\mathrm{Re}A\left(z,T,\delta\beta_{0}^{(p)}\right)= & \mathrm{Re}A\left(z,T,0\right)\cos\left\{ -\delta\beta_{0}^{(p)}z\right\} \label{eq:cor1}\\
 & -\mathrm{Im}A\left(z,T,0\right)\sin\left\{ -\delta\beta_{0}^{(p)}z\right\} ,\nonumber 
\end{align}

\begin{align}
\mathrm{Im}A\left(z,T,\delta\beta_{0}^{(p)}\right)= & \mathrm{Re}A\left(z,T,0\right)\cos\left\{ -\delta\beta_{0}^{(p)}z\right\} \label{eq:cor2}\\
 & +\mathrm{Im}A\left(z,T,0\right)\sin\left\{ -\delta\beta_{0}^{(p)}z\right\} .\nonumber 
\end{align}
Thus, PINNs can be applied to Eq. (\ref{eq:linear_multi}) with $\delta\beta_{0}^{(p)}=0$
to compute $A\left(z,T,0\right)$. Afterwards, the true values of
$\mathrm{Re}A(z,T,\delta\beta_{0}^{(p)})$ and $\mathrm{Im}A(z,T,\delta\beta_{0}^{(p)})$
can be restored through Eqs. (\ref{eq:cor1})-(\ref{eq:cor2}).

\section{Numerical results\label{sec:Numerical-results}}

We start our numerical analysis with the linear case, since it allows
us to determine the upper boundary of the propagation length which
can be processed by PINN. In fact, the complexity of the problem increases
with the propagation length. Indeed, as $z$ becomes larger than the
characteristic lengths $L_{D}$ and $L_{NL}$, the effects of dispersion
and nonlinearities in fibers become more pronounced resulting high
frequency oscillations of $\mathrm{Re}A$ and $\mathrm{Im}A$. According
to the F-principle \citep{fprinciple}, such cases require longer
time for training. Consequently, in the linear case, the maximum propagation
length is distributed an order of $L_{D}$. In the nonlinear case,
the maximum propagation length be at most of an order of $L_{D}$
or shorter depending on the $L_{NL}$. 

We consider linear and nonlinear cases for pulse energies $E=10$
nJ and $E=0.1$ nJ. The energy is equally divided between all modes.
We choose the energy $E=10$ nJ because of the small nonlinear length
$L_{NL}$, so we can observe nonlinear effects at small lengths of
fibers. On the other hand, the energy $E=0.1$ nJ is closer to realistic
values of energy, so we also consider this energy. The wavelength
for all cases is $\lambda=1030$ nm. The time window is 100 ps. Training
time is 2 hours unless otherwise specified. Table \ref{tab:parameters}
presents the dispersion coefficients for different lengths of fiber.
The values of $\widehat{\delta\beta_{0}^{(p)}}$are computed using
the scaling transformation (\ref{eq:bet})-(\ref{eq:bet1}).  

\begin{figure*}
\noindent \begin{centering}
\includegraphics[width=1\textwidth]{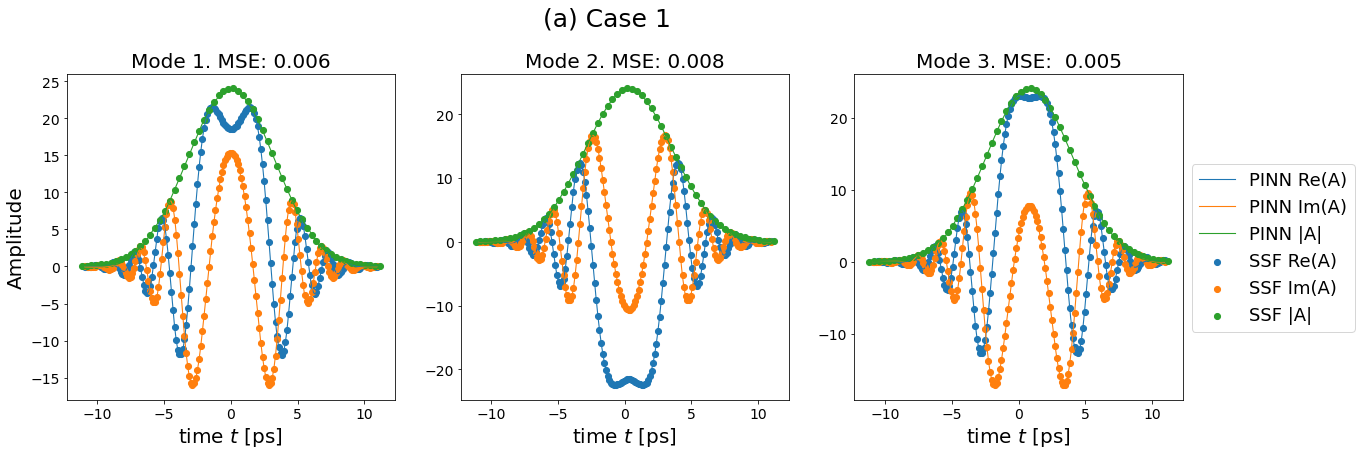}
\par\end{centering}
\begin{centering}
\includegraphics[width=1\textwidth]{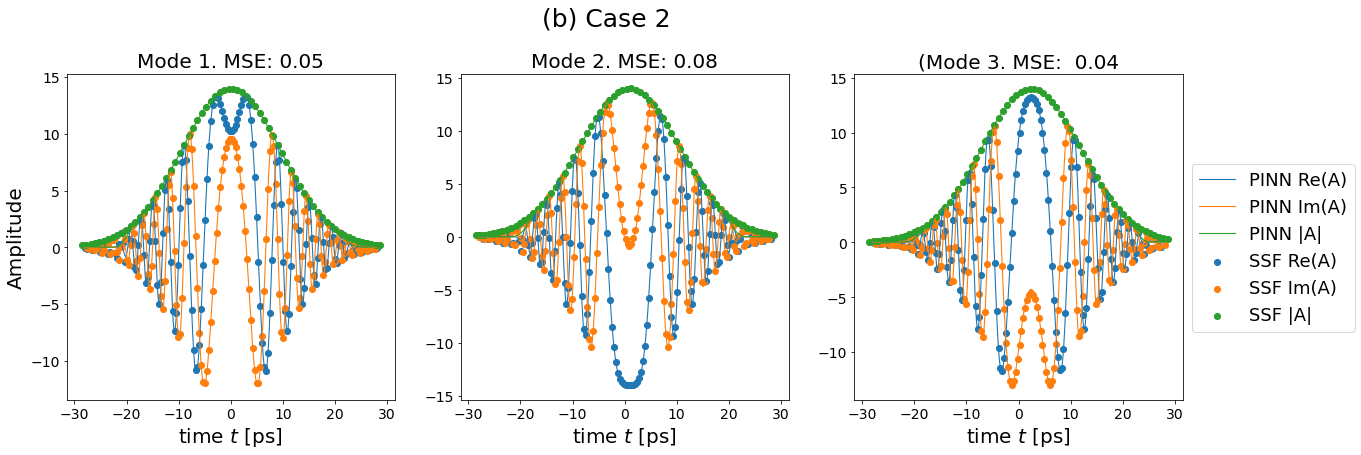}
\par\end{centering}
\caption{Case 1: three-mode linear case, $E=10$ nJ, $L=100$ m. Case 2: three-mode
linear case, $E=10$ nJ, $L=300$ m.}
\label{fig:results linear}
\end{figure*}

We examine the efficiency of scale transformation for a fiber of $L=100$
m length neglecting nonlinear effects (case 1). The SSF solution for
the problem with original values of $\delta\beta_{0}^{(p)}$and PINN
results with the scaled values of $\delta\beta_{0}^{(p)}$ are shown
in Fig. \ref{fig:results linear}(a). We see that the scaling of $\delta\beta_{0}^{(p)}$
allows to reproduce by PINN not only the absolute value of $U$, but
also its real and imaginary parts. Next, we increase the length of
the fiber up to $L=300$ m (case 2) and training time up to 2.5 hours.
The results are shown in Fig. \ref{fig:results linear}(b). The training
for 300 m requires slightly more time than for 100 m  as the solution
for 300 m case has much stronger oscillatory behavior (although the
increase of the training time is not significant). The agreement between
PINN and SSF is obtained keeping the MSE below 0.08.

\begin{table*}
\caption{Parameters of fiber and signal}

\label{tab:parameters}
\begin{centering}
\begin{tabular}{cccccc}
\hline 
\textnumero{} & $L$, m  & $p$ & $\widehat{\delta\beta_{0}^{(p)}}$, m$^{-1}$ & $\delta\beta_{1}^{(p)}$, ps/m & $\beta_{2}^{(p)}/2$, ps$^{2}$/m\tabularnewline
\hline 
\hline 
\multirow{3}{*}{1} & \multirow{3}{*}{100 / 300} & 1 & 0 & 0 & 0.01916410\tabularnewline
\cline{3-6} \cline{4-6} \cline{5-6} \cline{6-6} 
 &  & 2 & -0.09655858 & 0.00295601 & 0.01916082\tabularnewline
\cline{3-6} \cline{4-6} \cline{5-6} \cline{6-6} 
 &  & 3 & -0.00364598 & 0.00791849 & 0.01915536\tabularnewline
\hline 
\multirow{3}{*}{2} & \multirow{3}{*}{5} & 1 & 0 & 0 & 0.01916410\tabularnewline
\cline{3-6} \cline{4-6} \cline{5-6} \cline{6-6} 
 &  & 2 & -1.10186823 & 0.00295601 & 0.01916082\tabularnewline
\cline{3-6} \cline{4-6} \cline{5-6} \cline{6-6} 
 &  & 3 & -0.75762821 & 0.00791849 & 0.01915536\tabularnewline
\hline 
\end{tabular}
\par\end{centering}
\end{table*}

\textcolor{red}{}

\begin{figure*}
\begin{centering}
\includegraphics[width=1\textwidth]{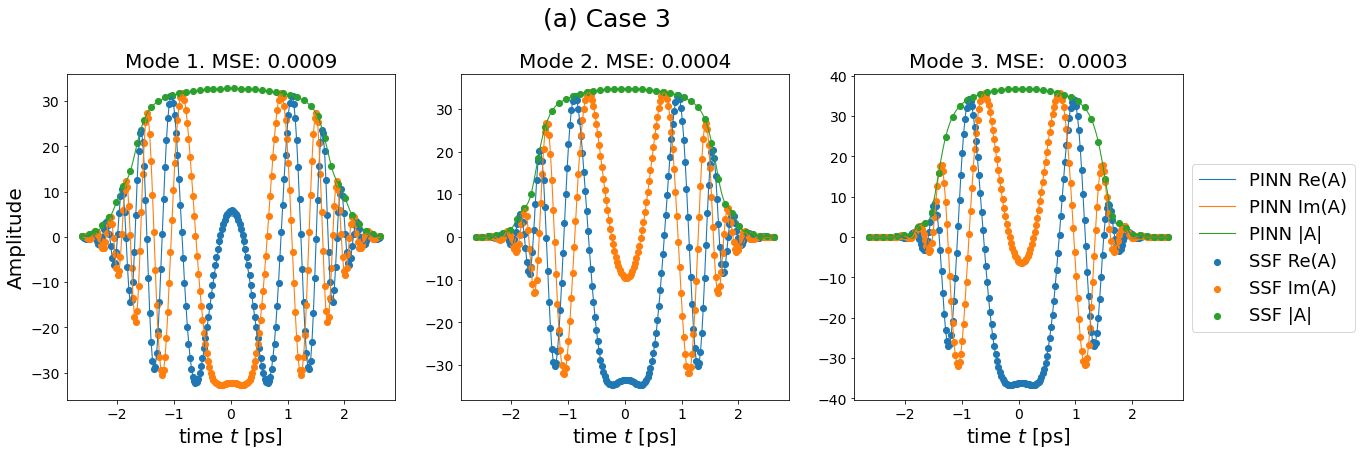}
\par\end{centering}
\begin{centering}
\includegraphics[width=1\textwidth]{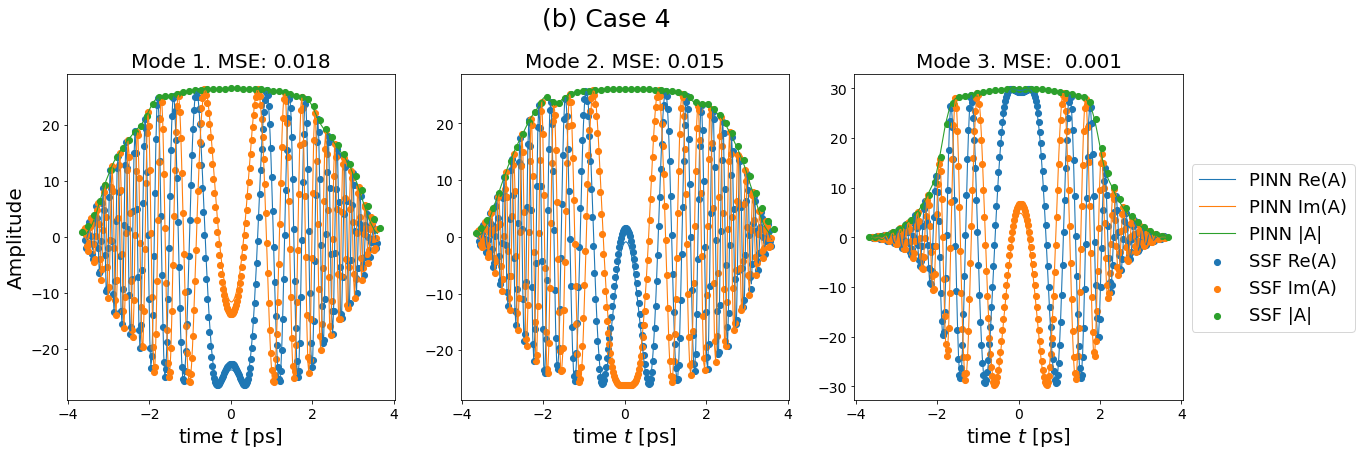}
\par\end{centering}
\caption{Case 3: three-mode nonlinear case with self-phase modulation, $E=10$
nJ, $L=5$ m. Case 4: three-mode nonlinear case with self- and cross-phase
modulations, $E=10$ nJ, $L=5$ m.}

\label{fig: results nonlinear 10 nJ}
\end{figure*}

Now we consider nonlinear cases. The results for the self-phase modulation
(case 3) with scaled values of $\beta_{0}^{(p)}$ are shown in Fig.
\ref{fig: results nonlinear 10 nJ}(a). In this example, the fiber
length is set to $L=5$ m, \textcolor{red}{} the pulse energy $E=10$
nJ. One can see that scale transformation provides robust PINN solution
in the nonlinear case. Note that for such high values of pulse energy,
the nonlinear length $L_{NL}$ is two orders of magnitude less than
the dispersion length $L_{D}$ and the former limits the maximum value
of the propagation length solvable by PINN. From Eq. (\ref{eq:nl-2})
we conclude that in the nonlinear case the maximum fiber length which
PINN can handle decreases when the pulse energy increases. 

Since in the pure self-phase modulation case (i.e.$\gamma_{C}^{(n)}=0$),
the equations for three modes are independent, we consider two configurations
of PINN. In the first one, a single PINN with six outputs is used
to compute three modes (as shown in Fig. \ref{pinn_scheme1}). In
the second configuration, three independent PINNs with two outputs
each are used for each mode. Our tests show that the first configuration
is just as efficient as the second one. Thus, the number of modes
does not impose principal difficulties for PINN.

Finally, we consider examples with both self-phase and cross-phase
modulation terms. The results for $E=10$ nJ and $L=5$ m (case 4)
are illustrated in Fig. \ref{fig: results nonlinear 10 nJ}(b). The
equations for three modes are coupled and PINN with six outputs is
used to solve a complete system of equations. The scaling procedure
of $\delta\beta_{0}^{(p)}$ provides robust results. Note, that for
this case the PINN training requires an increase in training time
up to 5 hours due to high frequency oscillations in real and imagery
parts of the solution. Essentially, at high energy $E=10$ nJ, the
nonlinear length $L_{NL}$ becomes several orders of magnitude smaller
than the dispersion length $L_{D}$, i.e. $L_{NL}\ll L_{D}$ . Thus,
at a distance of 5 m, we can clearly observe nonlinear effects. The
MMNLSE solution for longer fibers by PINN is a difficult problem and
requires substantial increase in the training time. 

\begin{figure*}
\begin{centering}
\includegraphics[width=1\textwidth]{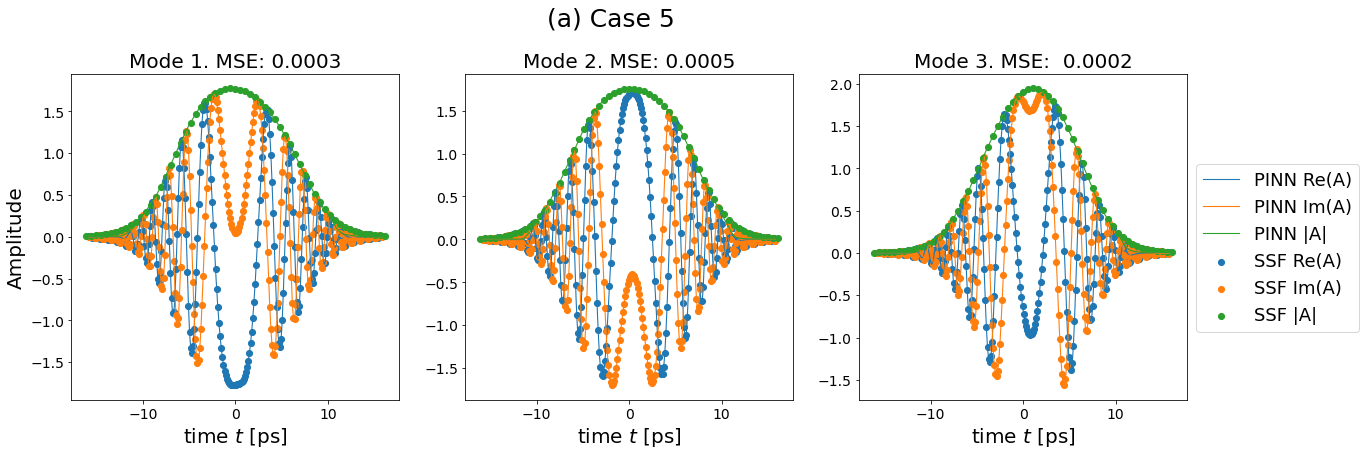}
\par\end{centering}
\begin{centering}
\includegraphics[width=1\textwidth]{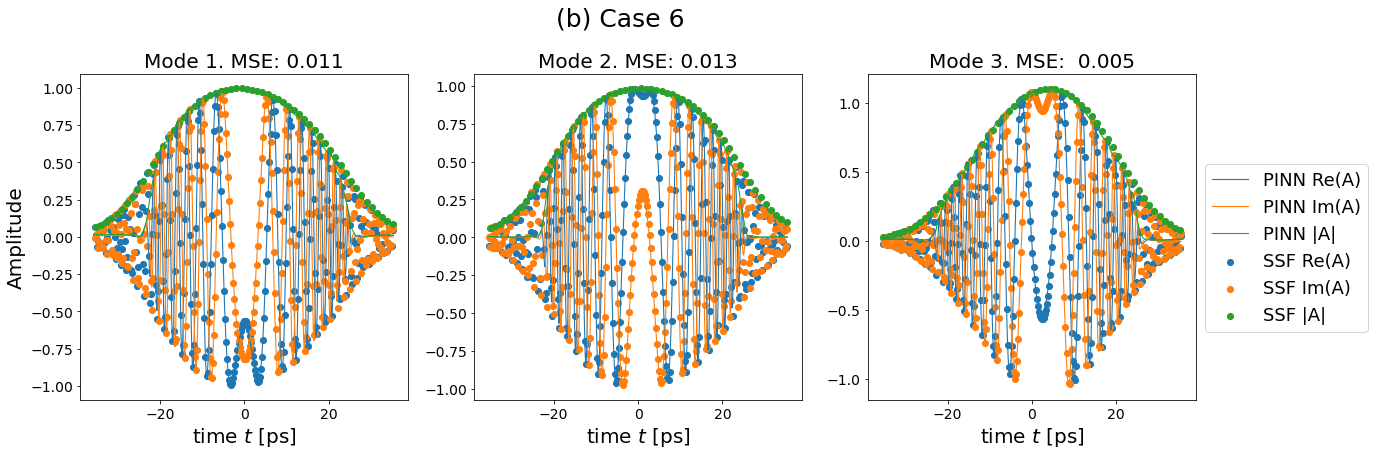}
\par\end{centering}
\caption{Case 5: three-mode nonlinear case with self- and cross-phase modulations,
$E=0.1$ nJ, $L=100$ m. Case 6: three-mode nonlinear case with self-
and cross-phase modulations, $E=0.1$ nJ, $L=300$ m.}

\label{fig: results nonlinear 01 nJ}
\end{figure*}

For pulse energy $E=0.1$ nJ (which is more realistic in applications),
the maximum lengths of fibers which PINN can handle increases. Fig.
\ref{fig: results nonlinear 01 nJ} shows solutions for three-mode
nonlinear cases with self- and cross-phase modulations for $E=0.1$
nJ, $L=100$ m (Fig. \ref{fig: results nonlinear 01 nJ}(a)) and $L=300$
m (Fig. \ref{fig: results nonlinear 01 nJ}(b)) (that are case 5 and
case 6, respectively). We can see that the default configuration of
PINN with 240000 sampling points (see Section \ref{sec:PINN-configuration})
yields good accuracy for case 5. By increasing the length of the fiber
up to 300 m, the number of points has to be increased up to 880000
dots and the training time up to 4 hours. Note that the increased
fiber length combined with nonlinear effects causes high-frequency
oscillations at the edges of the real and imaginary parts of the pulse,
which PINN cannot handle according to the F-principle. The results
show that the maximum length of fibers PINN can handle depends on
$L_{D}$ and $L_{NL}$. We found empirically that this length is about
$\mathrm{min}\left(50L_{D},50L_{NL}\right)$.\textcolor{red}{{} }

\textcolor{red}{}

\textcolor{red}{}

\section{Conclusion}

In this paper, the PINN has been applied to solve the MMNLSE. It was
shown that the zeroth order dispersion term in the MMNLSE causes high
frequency oscillations which add severe complexity for PINN. Even
if the absolute value of the temporal envelope goes not depend on
$\delta\beta_{0}^{(p)}$, PINN is not able to reproduce neither real
and imagery parts of the electric field temporal envelope, nor its
absolute value. It was shown that two factors impose computational
difficulty for PINN. The first one is the large difference of coefficients
at terms of MMNLSE. As a matter of fact, the optimizers used during
the PINN training might not capture the influence of terms with low
coefficients. This problem can be partially solved by using the frame
transformation and normalization, which however do not resolve the
difficulty associated with the $\delta\beta_{0}^{(p)}$-term. The
second factor is high oscillations in the solution.\textcolor{red}{{}
}In addition, the training time increases significantly according
to the F-principle \citep{fprinciple,roberts_yaida_hanin_2022}, which
states that a deep neural network tends to learn a target function
from low to high frequencies during the training. Note that the F-principle
limitation can be also used to explain why PINNs fail when the fiber
length becomes significantly larger than the dispersion and the nonlinear
lengths. 

In this study a novel scaling transformation to $\delta\beta_{0}^{(p)}$
for the MMNLSE solution by PINN has been proposed. A good agreement
is obtained between PINN and SSF. To our knowledge, it is the first
successful application of PINNs to the MMNLSE. Results are obtained
for fiber lengths up to 300 m for the pulse energy 10 nJ for linear
cases and 300 m for the pulse energy 0.1 nJ for nonlinear cases taking
into account self- and cross-phase modulations.\textcolor{red}{{} }We
note that including the cross-phase modulation term increases the
training time by two times as compared to the self-phase modulation
case. Besides, increasing the fiber length may require both an increase
in time and a multiple increase in the number of points. The maximum
fiber length which PINN can handle depends on the dispersion length
$L_{D}$ and nonlinear length $L_{NL}.$ Empirically we have found
that the maximum length can be roughly estimated as $\mathrm{min}\left(50L_{D},50L_{NL}\right)$.
Note that since $L_{NL}$depends on the pulse energy (see Eq. (\ref{eq:nl-2})),
for a given fiber length there is a maximum value of the pulse energy
PINN can handle.

In our future work we aim to include four wave mixing and Raman effect
in PINN as well as to apply the transfer learning technique \citep{Goswami2020,Chen2021}
which promises drastic computational savings compared to classical
approaches once the PINN has been trained. Also, another perspective
would be to study the performance of PINN for MMNLSE using a novel
network weight initialization scheme called Reptile described in \citep{Liu2022-ua}.

\section*{Appendix: solution for the single mode linear case}

Here, we summarize basic steps for solving Eq. (\ref{eq:linear_signl})
following \citep{Agrawal2021}. We consider a linear single mode equation
for $A(z,T)$

\begin{equation}
i\dfrac{\partial A}{\partial z}=\dfrac{\beta_{2}}{2}\dfrac{\partial^{2}A}{\partial T^{2}}.\label{eq:linear_signl-1}
\end{equation}
Performing Fourier transform we obtain

\begin{equation}
i\dfrac{\partial\hat{A}}{\partial z}=-\dfrac{1}{2}\beta_{2}\omega^{2}\hat{A}.\label{eq:fourier}
\end{equation}
Eq. (\ref{eq:fourier}) can be analytically integrated

\begin{equation}
\hat{A}\left(z,\omega\right)=\hat{A}(0,\omega)\exp\left\{ \dfrac{i}{2}\beta_{2}\omega^{2}z\right\} .
\end{equation}
The inverse Fourier transform reads as 

\begin{align}
A\left(z,T\right)=\dfrac{1}{2\pi} & \int_{-\infty}^{+\infty}\hat{A}\left(0,\omega\right)\label{eq:5}\\
 & \times\exp\left\{ \dfrac{i}{2}\beta_{2}\omega^{2}z-i\omega T\right\} d\omega,\nonumber 
\end{align}
where

\begin{equation}
\hat{A}\left(0,\omega\right)=\dfrac{1}{2\pi}\int_{-\infty}^{+\infty}U\left(0,T\right)\exp\left\{ \omega T\right\} dT.
\end{equation}
As the initial condition, we consider a Gaussian pulse given by 

\begin{equation}
A(0,T)=\exp\left\{ -\frac{T^{2}}{2T_{0}^{2}}\right\} ,\label{eq:bound}
\end{equation}
where $T_{0}$ is a half-width. Then applying Fourier transform to
Eq. (\ref{eq:bound}), we obtain

\begin{equation}
\hat{A}(0,\omega)=T_{0}\exp\left\{ -\frac{1}{2}T_{0}^{2}\omega^{2}\right\} .\label{eq:8}
\end{equation}
Substituting Eq. (\ref{eq:8}) into Eq. (\ref{eq:5}) yields 

\begin{align}
A(z,T)=\dfrac{1}{2\pi} & \int_{-\infty}^{+\infty}T_{0}\exp\left\{ -\frac{1}{2}T_{0}^{2}\omega^{2}\right\} \nonumber \\
 & \times\exp\left\{ \dfrac{i}{2}\beta_{2}\omega^{2}z-i\omega T\right\} d\omega.
\end{align}
Setting $a=$$\left(\dfrac{i}{2}\beta_{2}z-\frac{1}{2}T_{0}^{2}\right)$
and $b=$$iT$ and using the following expression 

\begin{equation}
\int_{-\infty}^{+\infty}\exp\left(a\omega^{2}+b\omega\right)d\omega=\sqrt{\dfrac{\pi}{-a}}\exp\left(-\dfrac{b^{2}}{4a}\right),
\end{equation}
we find that 

\begin{align}
A(z,T) & =\dfrac{T_{0}}{2\pi}\sqrt{\dfrac{\pi}{-\left(\dfrac{i}{2}\beta_{2}z-\frac{1}{2}T_{0}^{2}\right)}}\nonumber \\
 & \times\exp\left[-\dfrac{\left(iT\right)^{2}}{4\left(\dfrac{i}{2}\beta_{2}z-\frac{1}{2}T_{0}^{2}\right)}\right],
\end{align}
and after rearranging, we obtain Eq. (\ref{eq:lin}).


\end{document}